%% file: main.tex
\documentclass[sigconf ]{acmart}
\AtBeginDocument{%
  \providecommand\BibTeX{{%
    \normalfont B\kern-0.5em{\scshape i\kern-0.25em b}\kern-0.8em\TeX}}}

\copyrightyear{2024}
\acmYear{2024}
\setcopyright{rightsretained}
\acmConference[SETN 2024]{13th Conference on Artificial Intelligence}{September 11--13, 2024}{Piraeus, Greece}
\acmBooktitle{13th Conference on Artificial Intelligence (SETN 2024), September 11--13, 2024, Piraeus, Greece}\acmDOI{10.1145/3688671.3688770}
\acmISBN{979-8-4007-0982-1/24/09}
\acmPrice{15}
\graphicspath{{./images/}} 
\usepackage{stfloats}

\begin{document}

\title{The Large Language Model GreekLegalRoBERTa}


\author{Vasileios Saketos}
\authornotemark[1]
\email{saketos@kth.se}
\affiliation{%
  \institution{ KTH Royal Institute of Technology}
  \streetaddress{Brinellvägen 8, Stockholm}
  \city{Stockholm }
  \country{Sweden}
  %\postcode{SE-100 44 STOCKHOLM}
}

\author{Despina-Athanasia Pantazi}
\authornotemark[1]
\email{dpantazi@di.uoa.gr}
\affiliation{%
  \institution{National and Kapodistrian University of Athens}
  %\streetaddress{P.O. Box 1212}
  \city{Athens}
  \country{Greece}
}

\author{Manolis Koubarakis}
\authornotemark[1]
\email{koubarak@di.uoa.gr}
\affiliation{%
  \institution{National and Kapodistrian University of Athens}
  %\streetaddress{P.O. Box 1212}
  \city{Athens}
  \country{Greece}
}


\begin{abstract}

We develop four versions of GreekLegalRoBERTa, which are four large language models trained on Greek legal and nonlegal text. We show that our models surpass the performance of GreekLegalBERT, Greek- LegalBERT-v2, and GreekBERT in two tasks involving Greek legal documents: named entity recognition and multi-class legal topic classification. We view our work as a contribution to the study of domain-specific NLP tasks in low-resource languages, like Greek, using modern NLP techniques and methodologies.
\end{abstract}


%\ccsdesc[500]{Do Not Use This Code~Generate the Correct Terms for Your Paper}
%\ccsdesc[300]{Do Not Use This Code~Generate the Correct Terms for Your Paper}
%\ccsdesc{Do Not Use This Code~Generate the Correct Terms for Your Paper}
%\ccsdesc[100]{Do Not Use This Code~Generate the Correct Terms for Your Paper}

\begin{CCSXML}
<ccs2012>
<concept>
<concept_id>10010147.10010178.10010179.10010186</concept_id>
<concept_desc>Computing methodologies~Language resources</concept_desc>
<concept_significance>500</concept_significance>
</concept>
<concept>
<concept_id>10010147.10010178.10010179.10003352</concept_id>
<concept_desc>Computing methodologies~Information extraction</concept_desc>
<concept_significance>500</concept_significance>
</concept>
<concept>
<concept_id>10010147.10010178.10010179.10010184</concept_id>
<concept_desc>Computing methodologies~Lexical semantics</concept_desc>
<concept_significance>500</concept_significance>
</concept>
</ccs2012>
\end{CCSXML}

\ccsdesc[500]{Computing methodologies~Language resources}
\ccsdesc[500]{Computing methodologies~Information extraction}
\ccsdesc[500]{Computing methodologies~Lexical semantics}
\keywords{ Natural Language Processing, Pre-trained Language Models, Greek NLP Resources, Greek Legislation, Classification, Named Entity Recognition}

%% The following are not a requirement, delete if not using
%\received{20 February 2024}  %% inital submission date
%\received[revised]{12 March 2024} %% interim new draft
%\received[accepted]{5 June 2024}  %% publication version

\maketitle

\input{intro}
\input{related-work}
\input{experiments}

\input{conclusions}

\input{acknowledgments}

%\subsection{Do Not Remove Boilerplate Code}
% \begin{acks}
% Acknowledgements go here. Delete enclosing begin/end markers if there are no acknowledgements.
% \end{acks}

\bibliographystyle{ACM-Reference-Format}
\bibliography{main.bib}

\appendix

\section{Appendix}
\subsection{Pretraining hyperparameters}
In Table \ref{detailedComparison}, we provide an overview of the hyperparameters utilized during the pretraining phase for all the models.
\subsection{Finetuning hyperparameters}
In Table \ref{Finetuninghyperparameters}, we present the optimal hyperparameters for both the GreekLegalCode and GreekLegalNER tasks.
    \begin{table*}[b]
      \begin{center}
        \caption{Detailed comparison of the models' pretraining }
        \label{detailedComparison}
        \fontsize{8pt}{13.5pt}\selectfont\begin{tabular}{|c|c|c|c|c|c|} % <-- Changed to S here.
        \hline
         &\textbf{Training steps} & \textbf{Batch size} & \textbf{Vocabulary size} & \textbf{Size} & \textbf{Context}\\
        \hline
        \textbf{GreekBERT} &1M & 256& 35k& 29GB&nonlegal\\
        \hline
        \textbf{GreekLegalBERT} &1M & 256& 35k& 5GB&legal\\
        \hline
        \textbf{GreekLegalBERT-v2} &1M & 256& 35k& 8GB&legal\\
        \hline
        \textbf{GreekLegalRoBERTa-v1} &100K& 1024& 50k & 5GB &legal\\
          \hline
        \textbf{GreekLegalRoBERTa-v2} & 100K& 4096 & 50k & 5GB &legal\\
        \hline
        \textbf{GreekLegalRoBERTa-v3} &1M & 256& 50k& 8GB&legal\\
        \hline

        \textbf{GreekLegalRoBERTa-v4} &1M & 256& 50k& 37GB&legal and nonlegal\\

          \hline

        \end{tabular}
      \end{center}
    \end{table*}

    \begin{table*}[b]
      \begin{center}

        \caption{Best hyperparameters on finetuning the models for the experiments}
        \fontsize{8}{13.5pt}\selectfont 

\label{Finetuninghyperparameters}
        
        \begin{tabular}{|c|c|c|c|c|c|c|c|c|c|} % <-- Changed to S here.
          \hline
           &\multicolumn{2}{|c|}{\textbf{volume}}&\multicolumn{2}{|c|}{\textbf{chapter}}&\multicolumn{2}{|c|}{\textbf{subject}}&\multicolumn{3}{|c|}{\textbf{NER}}\\
           \hline
          \textbf{Model} & \textbf{learning rate} & \textbf{epochs} & \textbf{learning rate} & \textbf{epochs} & \textbf{learning rate} & \textbf{epochs}  & \textbf{learning rate} & \textbf{epochs}   & \textbf{batch size} \\
          \hline
          \textbf{GreekBERT} & 5e-5 & 3&3e-5&5&5e-5&15&5e-5&3& 8\\
          \hline
          \textbf{GreekLegalBERT}  &  3e-5 & 3&5e-5&5&3e-5&15&3e-5&3&8 \\
           \hline

          \textbf{GreekLegalBERT-v2}  &  3e-5 & 3&3e-5&8&3e-5&13&3e-5&3&8 \\
          \hline
          \textbf{GreekLegalRoBERTa-v1}  & 5e-5  & 6&3e-5&5&5e-5&15&5e-5&6&8\\
          \hline
          \textbf{GreekLegalRoBERTa-v2}  & 5e-5  & 6&5e-5&6&5e-5&14&5e-5&6&8 \\
          \hline
          \textbf{GreekLegalRoBERTa-v3}  & 5e-5  & 5& 5e-5 &13&5e-5&24&5e-5&3&8 \\
          \hline
          \textbf{GreekLegalRoBERTa-v4}  &3e-5  & 7& 5e-5 &21&5e-5 &18&3e-5 &7&8 \\
          \hline
        \end{tabular}
      \end{center}
    \end{table*}

\end{document}

%% file: intro.tex
\section{Introduction}
The success of the Transformer architecture \cite{DBLP:journals/corr/VaswaniSPUJGKP17} has led to the creation of many large language models \label{LLMs} (LLMs) and related benchmarks for the legal domain~\cite{DBLP:conf/emnlp/ChalkidisFMAA20,DBLP:conf/acl/ChalkidisJHBAKA22,DBLP:journals/corr/abs-2305-07507,DBLP:journals/corr/abs-2301-13126}. Of particular interest to us is the development of legal LLMs  and benchmarks for low-resource languages. 

For the Greek language in particular,~\cite{DBLP:conf/jurix/AngelidisCK18} studied the problem of named entity recognition (NER) \label{NER} in Greek legal documents using various kinds of RNN networks. The paper also developed the dataset GreekLegalNER\footnote{This name is not used in~\cite{DBLP:conf/jurix/AngelidisCK18}; it is the name~\cite{DBLP:journals/corr/abs-2301-13126} gave to this dataset (part of their LEXTREME benchmark), so we use the same name for consistency.}
which has been constructed using Greek legislation available in the platform Nomothesia~\cite{10.1007/978-3-319-58068-5_36}. Nomothesia is a platform that makes Greek legislation available on the Web as linked data using appropriate legal ontologies.

Later on,~\cite{papaloukas-etal-2021-multi} proposed GreekLegalBERT, a version of BERT trained on Greek legislation available in the platform Nomothesia~\cite{10.1007/978-3-319-58068-5_36}.  In~\cite{papaloukas-etal-2021-multi}, GreekLegalBERT was applied to the task of multi-class legal topic classification using the dataset GreekLegalCode consisting of 47k legal documents from Greek legislation.

In this work, we explore the use of RoBERTa~\cite{DBLP:journals/corr/abs-1907-11692} in the tasks of NER and multi-class legal topic classification for Greek legislation. RoBERTa is a version of BERT~\cite{devlin-etal-2019-bert} that has been obtained using a simpler training method which, nevertheless, leads to a significant gain in performance. The contributions of our paper are the following:
\begin{itemize}
    \item We develop four versions of GreekLegalRoBERTa. These four different LLMs were trained on the dataset available in the Nomothesia platform~\cite{10.1007/978-3-319-58068-5_36}, the Greek Parliament Proceedings \cite{NEURIPS2022_b96ce67b}, the Greek version of European Parliament Proceedings Parallel Corpus \cite{koehn-2005-europarl}, the Greek part of European Union legislation \cite{chalkidis-etal-2021-multieurlex}, Raptarchis \cite{papaloukas-etal-2021-multi} and OSCAR \cite{ortiz-suarez-etal-2020-monolingual}. We view our work as a contribution to the study of domain-specific NLP tasks in low-resource languages.
    \item We apply the trained models on the tasks of GreekLegalNER in Greek legal documents originally studied in~\cite{DBLP:conf/jurix/AngelidisCK18}, and multi-class legal topic for the GreekLegalCode dataset presented in~\cite{papaloukas-etal-2021-multi}. Our models managed to surpass the performance of all the previous models in micro and weighted average in GreekLegalNER and all the tasks of GreekLegalCode. More specifically, we managed to improve performance over the state-of-the-art results obtained by Greek- LegalBERT-v2 by 1.2 points in micro average and 1.4 points in weighted average within GreekLegalNER. Moreover, we managed to improve by 0.12 points in volume, 0.89 points in chapters and 0.67 points in subjects of GreekLegalCode. 
    \item We make our models, datasets, and code publicly available \footnote{ Our code is available at : \url{https://github.com/AI-team-UoA/GreekLegalRoBERTa} \newline
    Our models and datasets are available at: \url{https://huggingface.co/AI-team-UoA}
    }
    so that they can be used by the research community.
\end{itemize}

The organization of the rest of the paper is as follows. Section~\ref{sec:related-work} discusses related work. Section~\ref{sec:GreekLegalRoBERTa} analyzes the current state-of-the-art models and the different configurations we used to pretrain our models, while in Section~\ref{sec:experiments}, we apply the developed models to the tasks of NER  and multi-class legal topic classification. Section~\ref{sec:conclusions} concludes the paper by discussing limitations and future work.

%% file: related-work.tex
\section{Related Work}
\label{sec:related-work}
If one wants to solve an NLP task for a low resource language like Greek, one solution is to use a multilingual LLM. For Greek, both M-BERT and XLM-RoBERTa~\cite{DBLP:conf/acl/ConneauKGCWGGOZ20} will do for the task, since their training corpora includes Greek documents. The first monolingual language model to be proposed for the Greek language is GreekBERT~\cite{DBLP:conf/setn/KoutsikakisCMA20}. It has been applied to the tasks of part-of-speech (POS) tagging, NER and natural language inference (NLI), and it was shown to outperform M-BERT and XLM-RoBERTa on these tasks.

Nowadays, there has been a notable emphasis on the development and advancement of generative multilingual models. Some outstanding examples of these models are OPT\cite{DBLP:journals/corr/abs-2205-01068}, BLOOM \cite{DBLP:journals/corr/abs-2211-05100} and GPT3 \cite{DBLP:journals/corr/abs-2005-14165}, while the most recent and prominent are PaLM\cite{bi-etal-2020-palm}, Chinchilla\cite{DBLP:journals/corr/abs-2203-15556} and Llama\cite{touvron2023llama}. These models demonstrate significant performance in both natural and programming languages. While OPT, BLOOM, and Llama are open-source models trained on publicly available data, GPT-3, Chinchilla, and PaLM are closed source and trained on private datasets. Very recently, the generative LLM Meltemi has been developed for the Greek language~\footnote{\url{https://huggingface.co/ilsp/Meltemi-7B-v1.5}}. Nevertheless, considering that each model comprises hundreds of billions of parameters, the resources required to utilize these models are substantial. %Consequently, the following question appears. Are OPT and BLOOM truly available to everyone?

Moreover, in the generative question answering task, the responses of these models suffer from extensive hallucination. The term hallucination refers to the phenomenon where a model generates information that is incorrect, misleading, or entirely fabricated. Retrieval Augmented Generation (RAG) \cite{NEURIPS2020_6b493230} emerged to address this issue. RAG architectures consist of a retriever $p_{\eta}$   and a generator $p_{\theta}$. The generator is an encoder decoder architecture such as BART \cite{lewis-etal-2020-bart}, Llama \cite{touvron2023llama}, and  BLOOM \cite{DBLP:journals/corr/abs-2211-05100}. The retriever is an encoder architecture such as DPR \cite{Karpukhin2020DensePR}, E5 \cite{wang2024textembeddingsweaklysupervisedcontrastive} and Nomic \cite{Nussbaum2024NomicET}, and it is responsible for retrieving the top-k passages $(y_1,y_2,...,y_k)$ from a database given a query $x_i$. Then, the query along with the k passages are passed to the generator. Finally, the generator produces the response to the original query. Despite differences in the fine-tuning methods and architectures of the retrievers, they leverage the same objective. Their objective is to ensure that the \( \text{cosine\_similarity}(p_{\eta}(x_i), p_{\eta}(y_j)) \) accurately reflects the true relationship between the query \( x_i \) and the passage \( y_j \). To achieve this, they start from a pre-trained encoder model like RoBERTa and they fine-tune it according to this objective. Empirical results \cite{Shuster2021RetrievalAR} have shown that RAG reduces hallucination to a great extent.

Regarding the legal domain, the first language model proposed for Greek legislation is GreekLegalBERT~\cite{devlin-etal-2019-bert}, as we already mentioned in the introduction. It was applied to the task of multi-class legal topic classification for the dataset GreekLegalCode and it was shown to consistently outperform traditional machine learning algorithms (SVM and XG-BOOST), BiGRU-based methods and the multilingual models M-BERT and XLM-RoBERTa on this task.

With respect to the available legal benchmarks, the paper ~\cite{guha2023legalbench} presents LEGALBENCH, a collaboratively constructed legal reasoning benchmark which consists of 162 tasks covering six different types of legal reasoning. LEGALBENCH is important because legal professionals had a leading role in its construction. LEGALBENCH contains 112 legal binary classification tasks and 8 multi-class classification tasks. Recently,~\cite{DBLP:journals/corr/abs-2301-13126} proposed LEXTREME, a benchmark consisting of 11 legal datasets covering 24 languages and 18 tasks. They applied five popular encoder-based LLMs and found that the model size correlates with the performance on the benchmark in most cases. From the five models evaluated, XLM-RoBERTa is the most effective one. The datasets GreekLegalNER and GreekLegalCode studied in this paper are part of LEXTREME. What distinguishes our work from larger efforts such as LEXTREME is that by concentrating on a single language, we manage to develop language models that are more effective than the multilingual ones used in~\cite{DBLP:journals/corr/abs-2301-13126}. This paper reports on how we have achieved this using a version of RoBERTa trained on Greek legal documents.

%The paper ~\cite{guha2023legalbench} presents LEGALBENCH, a collaboratively constructed legal reasoning benchmark which consists of 162 tasks covering six different types of legal reasoning. LEGALBENCH is important because legal professionals had a leading role in its construction. Similar to GreekLegalCode, LEGALBENCH contains 112 Legal binary classification tasks and 8 multi-class classification tasks.

%% file: experiments.tex
\section{The GreekLegalRoBERTa models}
\label{sec:GreekLegalRoBERTa}

In this work, we focus on the use of RoBERTa and we develop four variations of our model GreekLegalRoBERTa. In the rest of the section, we provide the detailed description of the models in comparison to the existing state-of-the-art Greek models GreekBERT \cite{DBLP:conf/setn/KoutsikakisCMA20}, GreekLegalBERT \cite{athinaios-Legal-BERT} and GreekLegalBERT-v2 \cite{Greek-Legal-BERT-v2}. In Table \ref{Statistics on pretraining corpora}, we present the statistics of the pretraining corpora used in the various models.

\label{greekbert}
\textbf{GreekBERT \cite{DBLP:conf/setn/KoutsikakisCMA20}:} This model is a  monolingual version of BERT, trained solely on
    modern Greek, achieving state-of-the-art results in most of the Greek NLP tasks. BERT models are pretrained for 1M steps and batch size of 256. To speed up the pretraining process they used a sequence length of 128 for 90\% of the training steps. Then, for the remaining 10\% of steps, they used a sequence length of 512 to learn the positional embeddings. During pretraining the objective of BERT is to maximize the performance in Masked Language Modeling (MLM) and Next Sentence Prediction (NSP) \cite{DBLP:journals/corr/JerniteBS17}.
    GreekBERT was pretrained on 29GB of text from a corpus consisting of the Greek part of Wikipedia, the Greek part of the European Parliament Proceedings Parallel Corpus(Europarl) \cite{koehn-2005-europarl} and OSCAR \cite{ortiz-suarez-etal-2020-monolingual}. 

    \label{greeklegalbert}
    
    \textbf{GreekLegalBERT \cite{athinaios-Legal-BERT}:} \label{Bertv2} This model is a monolingual legal version of BERT trained on dataset accessible through the  Nomothesia~\cite{10.1007/978-3-319-58068-5_36} platform. The dataset consists of laws, announcements, and resolutions in the Greek language. The total size of the dataset is 5GB and it spans a chronological range from 1990 to 2017. Despite the smaller dataset, this model managed to exceed the performance of GreekBERT in GreekLegalCode and match its performance GreekLegalNER.
    %The pretraining procedure was the same as BERT's. The total number of parameters is 110M.

    \label{greeklegalbertv2}
    \textbf{GreekLegalBERT-v2 \cite{Greek-Legal-BERT-v2}:}  This model is a more recent version of GreekLegalBERT pretrained on the Nomothesia dataset, the Greek Parliament Proceedings (Greekparl) \cite{NEURIPS2022_b96ce67b}, Eurparl  \cite{koehn-2005-europarl}, the Greek part of European Union legislation (Eurlex) \cite{chalkidis-etal-2021-multieurlex} and Raptarchis \cite{papaloukas-etal-2021-multi}. The total size of the dataset sums up to 8GB. 
    %GreekLegalBERT-v2  managed to exceed the performa nce of GreekLegalBERT in GreekLegalCode and match its performance GreekLegalNER. 

    \label{GreekLegalRoBERTa-v1}
     \textbf{GreekLegalRoBERTa-v1: } This is the first model we train for the purposes of our work.  This model is pretrained on the dataset accessible through the  Nomothesia~\cite{10.1007/978-3-319-58068-5_36} platform. 
     
     The distinctive characteristics of our dataset render the preprocessing of the data essential. Due to the rather large chronological range of our data, a wide range of character encodings have been used. Windows-1253 and ISO 8859-7 were among the character encodings utilized. We therefore need to convert all identical characters into a unique representation. Moreover, we normalize our data using the normalization form compatibility decomposition (NFKD), because K normalizations are more effective in eliminating formatting distinctions.
    Additionally, we remove accents due to the possibility of words in Greek having the same letters but differing in their accents. 
    
    To encode our text we utilize Byte-Pair Encoding (BPE)\cite{sennrich-etal-2016-neural}, a hybrid between character and word level representations that allows handling the vocabularies common in natural language corpora. Instead of full words, BPE relies on subwords units, which are extracted by performing statistical analysis of the training corpus. To train our encoder, we utilize a vocabulary of size 50\,264. BPE training was done in the same dataset used to pretrain our model.
    
    In contrast to BERT, RoBERTa utilizes a sequence length of 512 throughout the entire pretraining process. This modification significantly increases the computational cost, as the attention mechanism's computational complexity grows quadratically with the sequence length. To avoid using the same mask for each training instance in every epoch in BERT, training data was duplicated 10 times so that each sequence is masked in 10 different ways over the 40 epochs of training. On the contrary, RoBERTa uses dynamic masking. In dynamic masking, the masking pattern is generated every time we feed a sequence to the model. In the original RoBERTa paper, it was concluded that dynamic masking performs slightly better than static masking. 
    By utilizing dynamic masking, we avoid duplication and consequently we reduce significantly  the memory requirements during training.
    Additionally, the authors removed the NSP objective as it was shown that using the NSP objective hurts the performance on downstream tasks. 
    The mixed floating point precision was used to train the original RoBERTa model. This technique increases stability during training while speeding up the training process.

    We pretrain for 100k steps and a batch size of 1024. This experiment solidifies the statement that pretraining RoBERTa leads to more prominent results even when we pretrain less compared to BERT. The model was pretrained using a single GPU. The total training duration of this model amounted to 30 days of uninterrupted training. %This model improved over GreekLegalBERT-v2 \ref{greeklegalbertv2} in GreekLegalCode by 0.08 points on the Volume and 0.05 points on the Chapter. Moreover, on the GreekLegalNER improved over GreekLegalBERT-v2 \ref{greeklegalbertv2} by 1 point on micro and weighted F1.

        \label{GreekLegalRoBERTa-v2}
     \textbf{GreekLegalRoBERTa-v2: } This model is identical to v1 except for the fact that we trained our model with a batch size of 4096 in a 4 GPU V100 cluster. In this model, our endeavor is to leverage the principle that within modern Natural Language Processing, more pretraining of a model correlates with enhanced performance. Moreover, training with large batches improves the perplexity for the
    masked language modeling objective, as well as the end-task accuracy. The total training duration of this model amounted to 40 days of uninterrupted training.
     %Our new  RoBERTa model outperforms all the previous models in GreekLegalCode metrics and in all but one in GreekLegalNER. 
     %More specifically GreekLegalRoBERTa-v2 improved over GreekLegalBERT-v2  by 0.14 points on the Volume 0.57 points on the Chapter and 0.01 on the Subject. Moreover on the GreekLegalNER improved over GreekLegalBERT-v2 by 0.4 points on micro and by 0.6 on weighted F1.

    \label{GreekLegalRoBERTa-v3}
     \textbf{GreekLegalRoBERTa-v3: } In our experiments discussed in section~\ref{sec:experiments}, it is evident that GreekLegalBERT-v2 exhibits improved performance compared to GreekLegalBERT. Our objective is to enhance the performance of our new model by leveraging the supplementary dataset employed in GreekLegalBERT-v2. We adopt the identical experimental setup as employed in GreekLegalRoBERTa-v2.
     %Our new model improved over GreekLegalRoBERTa-v2 \ref{greeklegalbertv2} by  0.32 points on the Chapter and 0.66 on the Subject. 

     \label{GreekLegalRoBERTa-v4}
     \textbf{GreekLegalRoBERTa-v4:} Last but not least, we utilize all the datasets we discussed in the previous models.
     In this model, we aim to ascertain whether incorporating both legal and non-legal contexts results in improved performance on legal tasks. We adopt the identical experimental setup as employed in GreekLegalRoBERTa-v2.
     %However, the model fails to demonstrate any discernible improvement in performance. This could be attributed to the substantial difference between the vocabulary employed in the Greek legal written context and that utilized in everyday life. As a consequence, our tokenizer inadequately represents the expanded vocabulary, leading to suboptimal performance.
     
     Having presented the models, we can now proceed to apply them and evaluate their performance.

\section{Experiments}
\label{sec:experiments}
We compare the performance of our newly introduced four GreekLegalRoBERTa models to the existing state-of-the-art Greek models GreekBERT, GreekLegalBERT and GreekLegalBERT-v2, against baselines on datasets for two core downstream tasks: NER and multi-class legal topic classification. In Tables \ref{NerReasults} and \ref{GLCexperiements}, we present our results. We observe that our models provide improvement over the originally reported performance of all the compared models.

\begin{table}[!htbp]
    \caption{Statistics of the pretraining corpora}
    \label{Statistics on pretraining corpora}
    \begin{tabular}
    {|p{2.7cm}|p{1.4cm}|p{1.5cm}|  }
    \hline
    \textbf{Corpus} & \hfil \textbf{Size (GB)} & \hfil \textbf{Context}\\
    \hline
    
    Nomothesia dataset & \hfil 5 & \hfil legal\\
    \hline
    Europarl & \hfil 0.38  & \hfil legal\\
    \hline
    Eurolex & \hfil 0.41 & \hfil legal \\
    \hline
    Greekparl & \hfil 2.7 & \hfil legal  \\
    \hline
    Raptrarchis & \hfil 0.22 & \hfil legal\\
    \hline
    Wikipedia & \hfil 1.73 & \hfil nonlegal\\
    \hline
    OSCAR & \hfil 27 & \hfil nonlegal\\
    \hline 
    \textbf{Total} & \hfil 37.03 & \hfil  -\\

    \hline
    \end{tabular}

    \end{table}

\label{greeklegalner}
\begin{table*}[!pt]

    \centering
      \begin{center}
        \caption{Results on the task of NER using F1 scores. To accommodate the results within the confines of the page, the utilization of the following acronyms is employed: Facility: F, Geopolitical Entity: GPE, Legislation Reference: LR, National Location: LN, Unknown Location: LU, Organization: ORG, Person: P, Public Document: PD.}
        \label{NerReasults}
        \fontsize{8pt}{14.2pt} {
            \begin{tabular}{|c|c|c|c|c|c|c|c|c|c|c|c|} % <-- Changed to S here.
          \hline
           & \textbf{F} & \textbf{GPE}  &
           \textbf{ LR }  &\textbf{ LN }&\textbf{LU }&\textbf{ORG}&\textbf{P}&\textbf{PD}& \textbf{micro}&\textbf{macro}&\textbf{weighted} \\

          \hline
          \textbf{GreekBERT}& 31 (3\%)&\textbf{75 (1\%)}&82 (0\%)&\textbf{88 (11\%)}&\textbf{73 (1\%)}&73 (1\%)&85 (1\%)&68 (1\%)&76.2&\textbf{71.8}&76.2\\
          \hline
          \textbf{GreekLegalBERT}  &29 (3\%)&74 (1\%)&82 (0\%)& 43 (16\%) &71 (2\%)&76 (1\%)&86 (1\%)&68 (2\%)&76&66&76\\
          \hline
          \textbf{GreekLegalBERT-v2}&31 (4\%)&\textbf{75 (1\%)}&82 (0\%)&62 (1\%)&72 (1\%)&76 (1\%)&87 (1\%)&67 (4\%)&76.4&69&76.2\\
          \hline
          \textbf{GreekLegalRoBERTa-v1} &30 (2\%)&\textbf{75 (0\%)}&82 (1\%)&55 (20\%)&71 (1\%)&\textbf{77 (1\%)}&87 (1\%)&68 (0\%)&76.8&68&76.8\\
          \hline
          \textbf{GreekLegalRoBERTa-v2}   &30 (2\%)&\textbf{75 (1\%)}&\textbf{ 83 (1\%)}&73 (19\%)&72 (1\%)&\textbf{77 (1\%)}&\textbf{88 (1\%)}&\textbf{70 (0\%)}&\textbf{77.6}&71&\textbf{77.6}\\
          \hline
          \textbf{GreekLegalRoBERTa-v3}&33 (1\%)&74 (1\%)&\textbf{83} (0\%)&69 (10\%)&71 (1\%)&76 (1\%)&\textbf{88 (1\%)}&\textbf{70 (1\%)}&76.8&70.5& 76.8\\
          \hline
          \textbf{GreekLegalRoBERTa-v4}&\textbf{35 (2\%)}&74 (1\%)&82 (1\%)&67 (2\%)&\textbf{73 (1\%)}&77 (1\%)&87 (1\%)&68 (1\%)&77&71&77\\
          \hline
          
        \end{tabular}}
        %\bottomrule

      % %
      % \footnote{To accommodate the results within the confines of the page, the utilization of the following acronyms is employed: Facility: F, Geopolitical Entity: GPE, Legislation Reference: LR, National Location: LN, Unknown Location: LU, Organization: ORG, Person: P, Public Document: PD.}

    \end{center}

    \end{table*}

\subsection{GreekLegalNER }

For the first downstream task, NER, we test the models against the benchmark GreekLegalNER, introduced by ~\cite{DBLP:conf/jurix/AngelidisCK18}.

\textbf{Dataset:} 
The dataset contains 254 daily issues of the Greek Government Gazette over the period 2000-2017. Every issue contains multiple legal acts. This dataset focuses on 7 entity types: legislation references, geopolitical entities, national locations, unknown locations, public locations, organizations, and facilities. The dataset is available in the Inside Outside Beginning (IOB) format, which is a common tagging format for tagging tokens for NER. It consists of 35\,411 instances and it is divided into 3 main parts: \textit{train} 67.5\%, \textit{ validation} 17.5\%, and \textit{test} 15\%.

\textbf{Evaluation:} We experiment with epochs from 1 to 20, batch size  8 and 16, learning rate 2e-5, 3e-5 and 5e-5. Table \ref{Finetuninghyperparameters} presents the hyperparameter combinations with which the models achieved their best results. Then, we perform 5 runs of finetuning and model evaluation per model using 5 different \textit{seeds}. For each model's performance on the \textit{test} set, we present the mean F1 score of every entity type, micro, macro and weighted F1, including the standard deviation of the 5 experiments in Table \ref{NerReasults}. In order to perform a comparative evaluation, we highlight the best F1 performance evaluation for each entity.

    \begin{table*}[!ht]
      \begin{center}
        \caption{Results on the task of multi-class legal topic classification}
        \label{GLCexperiements}
        \fontsize{8pt}{13.5pt}\selectfont\begin{tabular}{|c|c|c|c|c|c|c|c|c|c|} % <-- Changed to S here.
        \hline
         &\multicolumn{3}{|c|}{ \textbf{Volume}}&\multicolumn{3}{|c|}{ \textbf{Chapter}}& \multicolumn{3}{|c|}{\textbf{Subject}} \\
        \hline
         & Precision & Recall & F1 Score& Precision & Recall & F1 Score &  Precision & Recall & F1 Score \\
        \hline
        \textbf{GreekBERT} & 89.84 & 89.84& 89.84& 84.87& 84.87& 84.87& 80.59 & 80.59 &  80.59 \\
        \hline
        \textbf{GreekLegalBERT} & 90.51& 90.51& 90.51& 85.45& 85.45& 85.45& 81.43& 81.43& 81.43\\
        \hline
        \textbf{GreekLegalBERT-v2} & 91.02& 91.02 & 91.02& 85.72 &  85.72 &  85.72 & 82.71 &  82.71  &  82.71 \\
        \hline
        \textbf{GreekLegalRoBERTa-v1} & 91.10& 91.10& 91.10& 85.77& 85.77& 85.77& 82.29 & 82.29 & 82.29\\
          \hline
        \textbf{GreekLegalRoBERTa-v2} & \textbf{91.14} & \textbf{91.14} & \textbf{91.14} & 86.29 & 86.29 & 86.29 & 82.72 & 82.72 & 82.72\\
                  \hline
          \textbf{GreekLegalRoBERTa-v3}&89.04&89.04&89.04& \textbf{86.61}&\textbf{86.61}&\textbf{86.61}&\textbf{83.38}&\textbf{83.38}&\textbf{83.38}\\
          \hline
          \textbf{GreekLegalRoBERTa-v4}&91.02&91.02&91.02&86.31&86.31&86.31&82.54&82.54&82.54\\
          \hline
        \end{tabular}
      \end{center}
    \end{table*}

 Our model GreekLegalRoBERTa-v2 surpasses the performance of the state of the art GreekLegalBERT-v2 on micro F1 by 1.2 points and on weighted F1 by 1.4 points. Nevertheless, GreekBERT archives greater performance by 0.8 points than GreekLegalRoBERTa-v2. It is also evident that the RoBERTa models demonstrate superior performance compared to the BERT models trained on the same dataset.

Based on Table \ref{NerReasults},
GreekBERT achieves the highest score in 3 out of the 8 entity types, GreekLegalBERT in 0 out of 8, GreekLegalBERT-v2 in 1 out of 8, GreekLegalRoBERTa-v1 in 2 out of 8, GreekLegalRo-BERTa-v2 in 5 out of 8, GreekLegalRoberta-v3 in 3 out of 8, and GreekLegalRoberta-v4 in 2 out of 8. It is also clear that the models trained on non-legal contexts shown in Table \ref{detailedComparison} perform better in Facility entity type and National and Unknown Location types. This can occur because the datasets used during pretraining contain more instances of Facilities and Locations. On the other hand, models trained only on legal context tend to perform better in Person, Organization, and Public document entity types.

Furthermore, our anticipation was that the utilization of a larger dataset for training GreekLegalRoberta-v4 would result in a performance that significantly outperforms all the preceding models. However, contrary to our expectations, this was not the case.
This could be attributed to the substantial difference between the vocabulary employed in the Greek legal written context and that utilized in everyday life. As a consequence, our tokenizer inadequately represents the expanded vocabulary, leading to suboptimal performance.

\subsection{GreekLegalCode}
Lastly, for the second downstream task, multi-class legal topic classification, we conduct experiments using the GreekLegalCode dataset \cite{papaloukas-etal-2021-multi}. The GreekLegalCode paper introduces a new dataset of legal context and proves that using this dataset, GreekLegalBERT outperforms all the previous Greek and multilingual models. In Table \ref{GLCexperiements}, we show that our models outperform all the previous ones including the GreekLegalBERT versions.

\textbf{Dataset:}  The dataset is a thorough classification of the Greek legislation. It includes Laws, Royal and Presidential Decrees, Regulations, and Decisions, retrieved from the Official Government Gazette. The dataset is structured into thematic topics, making the data ideal for multi-label classification. It consists of 47 legislative \textit{volumes} and each \textit{volume} corresponds to a main thematic topic. Each \textit{volume} is divided into thematic subcategories which are called \textit{chapters} and subsequently, each \textit{chapter} breaks down into \textit{subjects}. The total number of \textit{chapters} is 389 while the total number of \textit{subjects} is 2285.

\textbf{Evaluation:} As proposed in the original paper\cite{papaloukas-etal-2021-multi}, we perform grid-search over the
 core hyper-parameters. More specifically, we experiment with epochs from 1 to 20 and learning rate 1e-5, 2e-5, 3e-5, and 5e-5.
We present the best hyperparameter configuration in Table \ref{Finetuninghyperparameters}. 
The experimental setup is the same as in GreekLegalNER, analyzed in the previous subsection.
We present the mean of \textit{micro} F1, precision, and recall of the performance evaluation in Table \ref{GLCexperiements}.
We start from the established benchmark obtained by GreekLegalBERT-v2 which is 91.02 for Volume, 85.72 for Chapter, and 82.71 for Subject.
Our model GreekLegalRoBERTa-v2 managed to achieve greater performance by 0.12 points on Volume. Moreover  GreekLegalRoBERTa-v3 surpasses the performance of GreekLegalBERT-v2 by 0.89 on chapter and 0.67 on subject.

%% file: conclusions.tex
\section{Conclusion and Future work}
\label{sec:conclusions}
In this work, we introduce four new language models pretrained in legal and nonlegal text. We have attained state-of-the-art results across all the tasks of the GreekLegalCode dataset.
Moreover, our models demonstrate superior performance compared to all the other models across all metrics, with the exception of the macro average in GreekLegalNER. Among the two conducted experiments, GreekLegalNER emerges as the more challenging task. One reason for this can be that the models have difficulty classifying correctly the two different types of location entities. 

Finally, for future work, it would be interesting to see how well the models perform if we combine the two different location categories. Moreover, we aim to develop more Natural Language Processing models in modern Greek and legal contexts, as we believe it is crucial, especially in low-resource languages like Greek. By providing NLP models in specific and non-specific domains not only do we create opportunities and tools for future research, but we also help the industry to provide more reliable products to everyone.

Furthermore, nowadays MergeKit \cite{goddard2024arcees} provides the opportunity to merge different NLP models of the same architectures in any hardware. By combining two NLP models, the new model can utilize each other's strengths without additional pretraining, offering a promising avenue for enhancing a model's performance. Developing additional NLP models for specific and non specific context is essential to fully capitalize on this opportunity.

Last but not least, our model can be used as a foundation for developing high-quality retrievers like  DPR \cite{Karpukhin2020DensePR}, and E5 \cite{wang2024textembeddingsweaklysupervisedcontrastive}, specifically for Greek legislation. By training encoder decoder architecture like  BART \cite{lewis-etal-2020-bart}, or Llama \cite{touvron2023llama} in Greek legal text, we can establish a highly effective architecture for Retrieval Augmented Generation (RAG) tailored to the Greek legislation.

%% file: acknowledgments.tex
\begin{acks}
We would like to express our sincere gratitude to the Cyprus Research Institute for providing us with access to Cyclone. Cyclone is a cluster which consists of 16 40-core compute nodes and 4 Nvidia V100 GPUs each. This work would not have been possible without this contribution.
\end{acks}